\documentclass[10pt,twocolumn,letterpaper]{article}

\usepackage[pagenumbers]{cvpr} 

\definecolor{cvprblue}{rgb}{0.21,0.49,0.74}
\usepackage[pagebackref,breaklinks,colorlinks,allcolors=cvprblue]{hyperref}

\usepackage{bm}
\usepackage{amsmath}
\usepackage{amssymb}
\usepackage{amsthm}

\usepackage{multirow}
\usepackage{multicol}

\usepackage{adjustbox}
\usepackage{booktabs}
\usepackage{threeparttable}
\usepackage{tabularx}

\usepackage{algorithm}
\usepackage{algorithmicx}
\usepackage[noend]{algpseudocode}
\usepackage{bbding}

\usepackage{pifont}
\usepackage{mathtools}
\usepackage{thmtools}
\usepackage{thm-restate}

\newcommand{\methodname}{PhyCustom}
\newcommand{\task}{physical customization}

\usepackage[capitalize]{cleveref}
\Crefname{section}{Sec.}{Secs.}
\Crefname{section}{Section}{Sections}
\Crefname{table}{Table}{Tables}
\Crefname{table}{Tab.}{Tabs.}

\newcommand{\cB}{\mathbf{c}}

\newcommand{\pB}{\mathbf{p}}
\newcommand{\xB}{\mathbf{x}}

\newcommand{\zB}{\mathbf{z}}

\newcommand{\Dcal}{\mathcal{D}}
\newcommand{\Ecal}{\mathcal{E}}

\newcommand{\Lcal}{\mathcal{L}}
\newcommand{\Ncal}{\mathcal{N}}

\newcommand{\Ucal}{\mathcal{U}}

\newcommand{\Ebb}{\mathbb{E}}
\newcommand{\Rbb}{\mathbb{R}}

\def\paperID{5431} 
\def\confName{CVPR}
\def\confYear{2026}

\title{PhyCustom: Towards Realistic Physical Customization in Text-to-Image Generation}

\author{
    Fan Wu\textsuperscript{1} \quad
    Cheng Chen\textsuperscript{1} \quad
    Zhoujie Fu\textsuperscript{1} \quad
    Jiacheng Wei\textsuperscript{1} \quad
    Yi Xu\textsuperscript{2} \quad
    Deheng Ye\textsuperscript{3} \quad
    Guosheng Lin\textsuperscript{1}\thanks{Corresponding author.} \\
    \textsuperscript{1}Nanyang Technological University \quad \textsuperscript{2}Goertek Alpha Labs \quad
    \textsuperscript{3}Tencent \\
    {\tt\small fan011@e.ntu.edu.sg \quad gslin@ntu.edu.sg} \\
    {\small \textcolor{magenta}{\texttt{https://github.com/wufan-cse/PhyCustom}}}
}

\begin{document}
\twocolumn[{
\renewcommand\twocolumn[1][]{#1}
\maketitle
\begin{center}
    \captionsetup{type=figure}
    \includegraphics[width=0.9\linewidth]{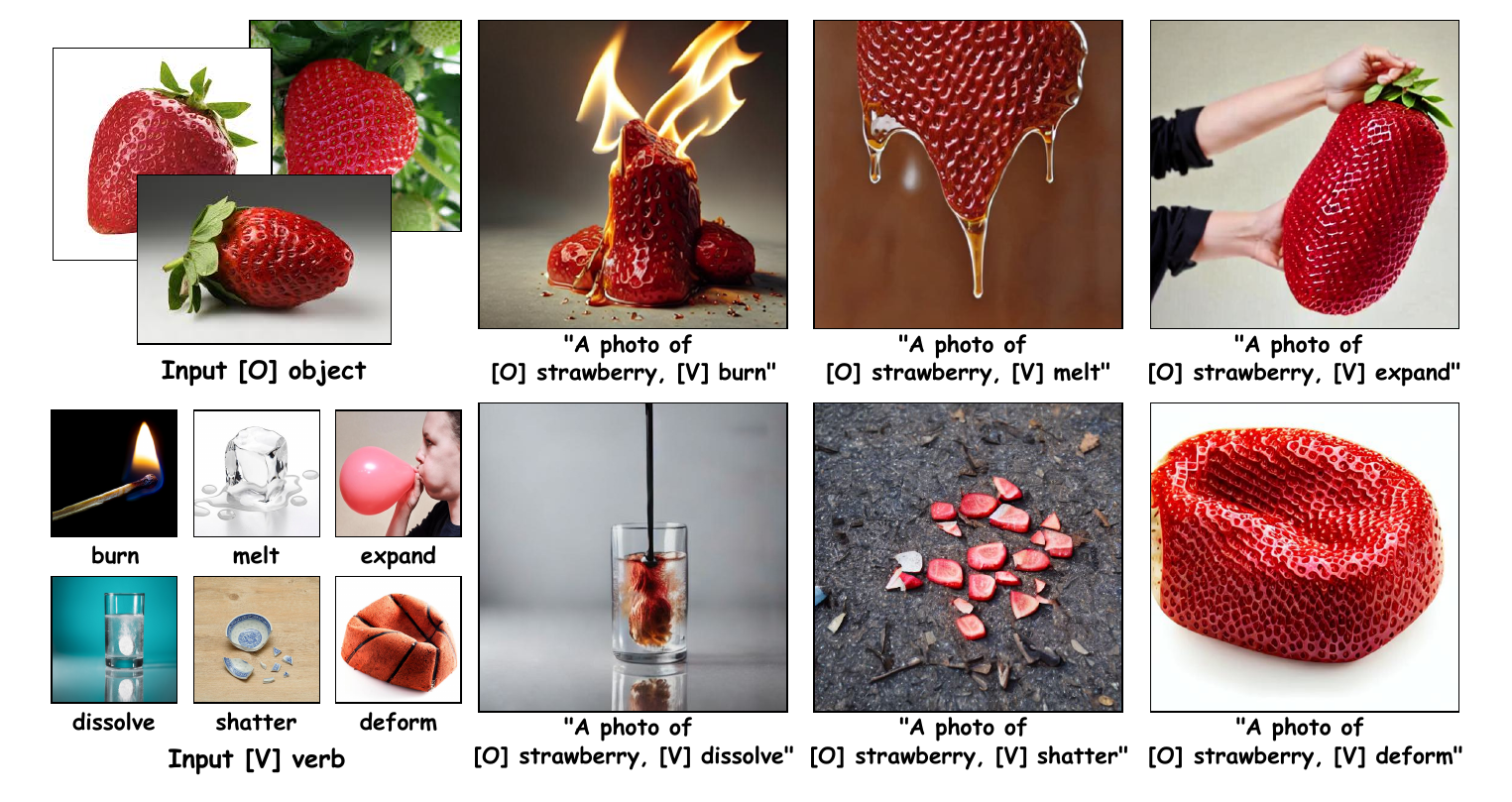}
    \captionof{figure}{\textbf{\methodname's ability.} We present some results of our proposed~\methodname~to showcase its ability in performing realistic physical transformations on common objects to generate novel concepts using a limited number (2$\sim$5) of provided images and corresponding texts.
    }
    \label{fig:teaser}
\end{center}
}]

\begin{abstract}
Recent diffusion-based text-to-image customization methods have achieved significant success in understanding concrete concepts to control generation processes, such as styles and shapes.
However, few efforts dive into the realistic yet challenging customization of physical concepts.
The core limitation of current methods arises from the absence of explicitly introducing physical knowledge during training. 
Even when physics-related words appear in the input text prompts, our experiments consistently demonstrate that these methods fail to accurately reflect the corresponding physical properties in the generated results.
In this paper, we propose~\methodname, a fine-tuning framework comprising two novel regularization losses to activate diffusion model to perform physical customization.
Specifically, the proposed isometric loss aims at activating diffusion models to learn physical concepts while decouple loss helps to eliminate the mixture learning of independent concepts.
Experiments are conducted on a diverse dataset and our benchmark results demonstrate that~\methodname~outperforms previous state-of-the-art and popular methods in terms of physical customization quantitatively and qualitatively.

\end{abstract}

\section{Introduction}
\begin{figure}[t]
    \centering
    \includegraphics[width=\linewidth]{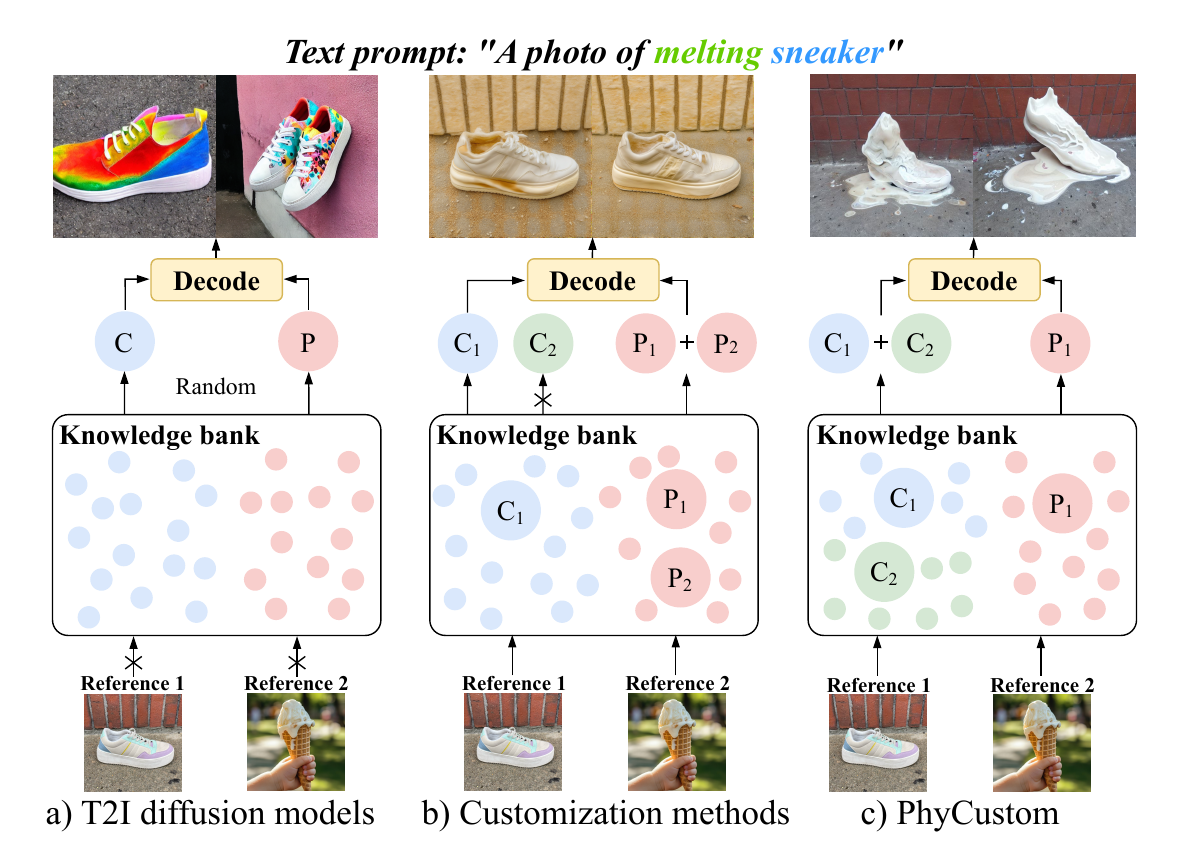}
    \caption{\textbf{Comparison of different methods.} 
    \textbf{a)} Given the text prompt, T2I diffusion models randomly select a pattern (e.g. style) associated with the object concept.
    \textbf{b)} After being fine-tuned on reference images, customization methods learn the patterns and then merge them by preserving the shape from reference 1 and the style from reference 2, however, fail to learn the physical concept.
    \textbf{c)}~\methodname~learns the physical concept and perform a concept-level merge to generate desired results.
    }
    \label{fig:intro_comparison}
\end{figure}

Generative models have seen remarkable advancements, especially with the rise of text-to-image(T2I) diffusion models~\cite{Rombach_2022_CVPR, balaji2022ediff, nichol2021glide, saharia2022photorealistic, yu2022scaling}, which have established a high standard for creating realistic images across diverse domains. 
Recently, numerous image customization works~\cite{alaluf2023neural, cao2023masactrl, chen2024anydoor, gal2023encoder, kawar2023imagic, tewel2023key, chen2023photoverse, ding2024freecustom, li2023blip, motamed2023lego, kumari2023multi, zhang2023adding} have devoted to control the diffusion process to generate fantastic and desired results.
However, most of them mainly concentrate on the customization of concrete concepts, which are perceivable like shapes and styles, and few of them seek to investigate the challenging physical concepts, which are descriptive terms and hard to perceived directly.
For example, we can see ``a rock'' (a concrete object with a specific shape and style), but its physical attributes like density are not perceivable. 
Furthermore, existing customization methods largely operate by merging visual patterns, where they aim at generating images with a natural combination of the reference patterns as shown in~\Cref{fig:intro_comparison}.
While effective for customizing concrete concepts, this pattern-level merging falls short for physical customization, which requires a higher-level, conceptual merging beyond visually perceivable patterns.

In this paper, we target at~\task, where we aim at activating T2I diffusion models to perform physical transformations on the target object given limited reference images.
Physical customization primarily comprises two challenges: the learning of physical concepts and the combination of different concepts.
The first challenge involves how to activate the diffusion models to extract the physical knowledge from cross-object data.
Unlike learning the object concept (e.g., ``teapot''), where images consistently depict the same concept, in our setting, physical concepts are implicitly depicted by cross-object data. 
For example, the concept of ``burn'' is depicted by images of burning wood, a match, and a house, each paired with corresponding texts.
The second challenge is how to decouple the learning of different concepts and achieve a natural combination of object and physical concepts. 
T2I diffusion models often struggle to learn multiple concepts independently, which leads to pattern leaking—where the generated object unintentionally inherits styles or shapes from one concept while applying another. 
Therefore, ensuring that different concepts can be combined independently is crucial for successful physical customization.

To address the challenges mentioned above, we propose a brand-new fine-tuning framework, namely~\methodname, equipping diffusion models to learn the physical concepts and perform independent concept merging as shown in~\Cref{fig:intro_comparison}.
The central innovation of~\methodname~is the introduction of two specialized regularization loss functions. 
The first one is the isometric loss, which aims at finding physics-related embedding from the given cross-object data to activate T2I diffusion models to learn the physical concepts.
The second one is the concept decouple loss, which regularizes the gradients of different concepts to be orthogonal, in a way to ensure the learning of different concepts independently.
These novel losses together empower T2I diffusion models to perform a wide range of physical customizations, as shown in~\Cref{fig:teaser}.

In our experiments, we evaluate~\methodname~on a newly constructed diverse dataset.
Experimental results show that our proposed method significantly outperforms popular and state-of-the-art (SOTA) baselines, both quantitatively and qualitatively.
Our contributions can be summarized as:
\begin{itemize}
    \item We introduce the task of physical customization and are the first to explore a wide range of physical concepts on T2I generation.
    \item We propose a novel framework equipped with two well-designed loss functions, enabling diffusion models in freely controlling physical customization.
    \item We conduct extensive experiments across various evaluation metrics, demonstrating that our proposed method outperforms SOTA baselines.
\end{itemize}

\section{Related Works}

\subsection{Text-to-Image Generation}
Numerous works for text-to-image generation have exhibited superior performance and can be classified into two categories.
The works from the first category are mainly based on Generative Adversarial Networks (GANs), for example, AttnGAN~\cite{xu2018attngan}, StackGAN~\cite{zhang2017stackgan}, StackGAN++~\cite{zhang2018stackgan++}, Mirrorgan~\cite{qiao2019mirrorgan}, Df-gan~\cite{tao2022df} and Dae-gan~\cite{ruan2021dae}. 
In recent years, with the advance of computational devices and large-scale pre-training datasets, text-to-image diffusion models~\cite{nichol2021glide, saharia2022photorealistic, Rombach_2022_CVPR, ramesh2022hierarchical} have shown significant success in generating high-quality images conditioned on text descriptions and there is an eruption of diffusion-based models, for example, Stable Diffusion (SD)~\cite{Rombach_2022_CVPR}, Imagen~\cite{saharia2022photorealistic}, DALLE2~\cite{ramesh2022hierarchical}, and GLIDE~\cite{nichol2021glide}.
These works consistently deliver high-resolution images with specific text prompt inputs.
However, diffusion-based models struggle to understand physical concepts because these concepts often co-occur with specific objects, causing models to mistakenly interpret them as parts of the object's appearance or style.

\subsection{Image Customization}
Image customization~\cite{gal2022image, alaluf2023neural, cao2023masactrl, chen2024anydoor, gal2023encoder, kawar2023imagic, tewel2023key, chen2023photoverse, zhu2024multibooth, li2024blip, ding2024freecustom, zhao2024uni} aims to generate images of interest, given a specific condition, which takes text or image as an input prompt.
A pioneering approach, DreamBooth~\cite{ruiz2023dreambooth}, fine-tunes the full model weights of a text-to-image diffusion model to bind a unique identifier with the subject of interest.
Custom Diffusion~\cite{kumari2023multi} and Perfusion~\cite{tewel2023key}, address multi-concept composition by modifying the model’s architecture or embedding process to support multiple subjects in a single image.
FreeCustom~\cite{ding2024freecustom} introduces a tuning-free approach that leverages a Multi-Reference Self-Attention (MRSA) mechanism and weighted masks.
ControlNet~\cite{zhang2023adding} introduces the first adapter specifically designed for task-specific input conditions in text-to-image diffusion models, enabling fine-grained adjustments based on conditions like edge maps.
IP-adapter~\cite{ye2023ip} proposes the decoupled cross-attention mechanism, which independently processes text and image features, enabling effective image prompts without modifying the original diffusion model’s parameters.
B-LoRA~\cite{frenkel2024implicit} finds that jointly learning the LoRA weights achieves style-content separation for image customization.
ZipLora~\cite{shah2024ziplora} proposes to train style and subject LoRAs individually and then merge to generate desired combination.
T2I-Adapter~\cite{mou2024t2i} includes a style-specific adapter that enables generation aligned with reference images.
LEGO~\cite{motamed2023lego} addresses the problem beyond customizing object appearances, but it falls short in considering various physical concepts compared to~\methodname.
Orthogonal Adaptation~\cite{po2024orthogonal} introduces orthogonal regularization on LoRA weights to enable modular customization. However, restricting representations to a small subset of orthogonal features limits its ability to handle physical customization, as physical concepts are inherently more complex and harder to learn, which may require richer features for learning, compared with clearly defined concepts.

\begin{figure*}[t]
    \centering
    \includegraphics[width=\linewidth]{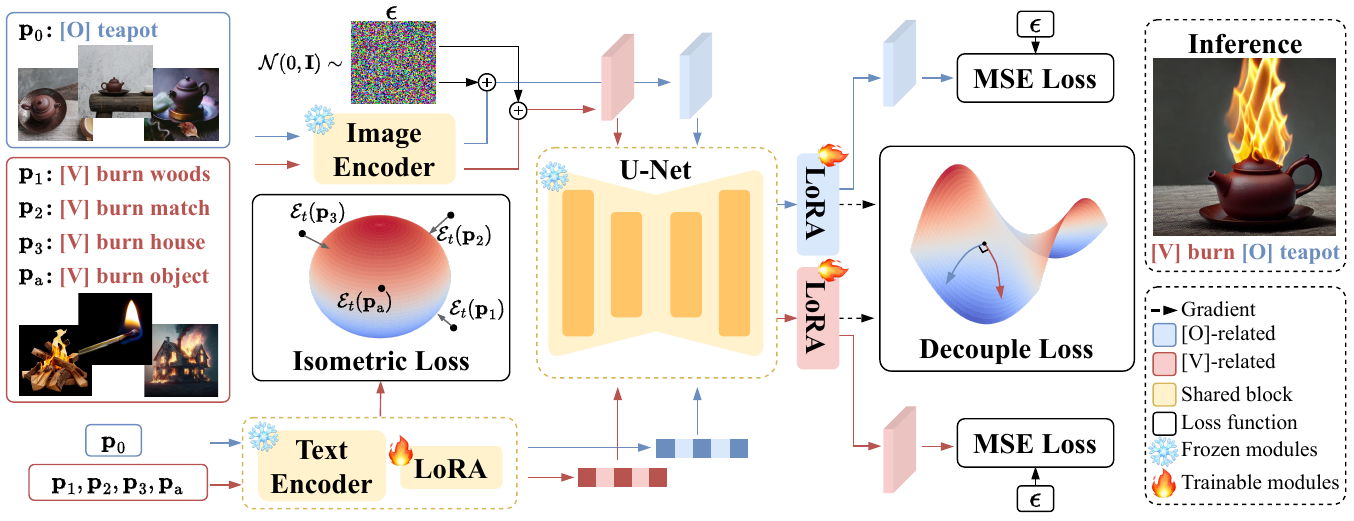}
    \caption{\textbf{Overview of~\methodname.} Given two sets of images and their corresponding prompts $\{\pB_0, \pB_1, \pB_2, \pB_3\}$, we fine-tune the diffusion model in a single stage with three losses.
    (1) The diffusion model with trainable LoRA modules is fine-tuned by MSE loss according to~\Cref{eq:diffusion_training_loss}, while the original parameters are frozen.
    (2) The isometric loss calculated by~\Cref{eq:cross_context_loss} aims at fine-tuning the text encoder to find a subspace where the distances between $\pB_i, i \in \{1, 2, 3\}$ and $\pB_\text{a}$ are equal, thus, the text encoder is able to learn the invariant text embedding, which is physics-related.
    (3) The decouple loss calculated by~\Cref{eq:decouple_loss} aims at decoupling the learning of different concept features by regularizing the gradient descents along two orthogonal directions.}
    \label{fig:main_flow}
\end{figure*}

\section{Methodology}
In this section, we first review relevant techniques in~\Cref{subsec:preliminaries}. 
We then introduce our proposed isometric loss for learning physical concepts in~\Cref{subsec:isometric_regularization}. 
Finally, we present our decoupling loss in~\Cref{subsec:concepts_decoupling}, which aims to prevent the entangled learning of independent concepts to perform physical customization.

\subsection{Preliminaries}
\label{subsec:preliminaries}

\paragraph{Diffusion models.} 
Stable diffusion~\cite{Rombach_2022_CVPR} are generative models that synthesize data samples by progressively transforming Gaussian noise into a structured output through iterative denoising steps. 
Given an input image $\xB$, it is encoded into a latent representation $\zB$ via a variational autoencoder (VAE)~\cite{kingma2013auto} encoder $\Ecal$, yielding $\zB = \Ecal (\xB)$. 
The diffusion process then injects varying levels of noise $\epsilon$ into $\zB$, resulting in $\zB_t$, where $t$ represents discrete time steps. 
The U-Net~\cite{ronneberger2015u} $\Ucal$ with parameters $\theta$ is trained to predict the added noise within the noisy latent $\zB_t$ based on a given text condition $\cB$. 
Training typically uses a mean-squared error loss:
\begin{equation}
\label{eq:diffusion_training_loss}
    \Lcal = \Ebb_{\Ecal(\xB), \epsilon \sim \Ncal(0,1), t}\left[\| \epsilon - \Ucal_\theta(\zB_t, \cB) \|_2^2 \right] ,
\end{equation}
where $\Ncal(0,1)$ is the standard Gaussian distribution with 0 mean and 1 variance.
In this work, we utilize Stable Diffusion~\cite{Rombach_2022_CVPR} as our base model.


\paragraph{LoRA fine-tuning.} 
LoRA~\cite{hu2021lora} is an effective approach to optimize large models by freezing the pre-trained model weights and introducing trainable, low-rank matrices. 
For a weight matrix $\bm{W} \in \mathbb{R}^{n \times m}$ in the diffusion model, LoRA adds a trainable low-rank matrix pair, updating $\bm{W}$ to $\bm{W}' = \bm{W} + \bm{B} \cdot \bm{A}$ where $\bm{B} \in \Rbb^{n \times r}$ and $\bm{A} \in \mathbb{R}^{r \times m}$ with $r \ll \min(n, m)$. 
In multi-LoRA composition, multiple $k$ LoRA modules are merged into the diffusion model using a linear combination $\bm{W}' = \bm{W} + \sum_{i=1}^{k} w_i \cdot \bm{B}_i \cdot \bm{A}_i$, where $w_i$ is a pre-defined weight for each LoRA network.
In this work, we train Low-Rank Adaptation (LoRA)~\cite{hu2021lora} modules instead of the whole diffusion model to accelerate our fine-tuning process.

\subsection{Isometric Regularization}
\label{subsec:isometric_regularization}

We propose a novel regularization method, namely isometric regularization.
A key characteristic contributes to the failure of diffusion models in physical customization is that physical concepts are often depicted along with a specific object.
As a result, diffusion models tend to treat physical concepts as part of object's appearance, despite being pre-trained on large-scale, diverse datasets that contain implicit physics knowledge.
The goal of isometric regularization is to find the physics-related text embedding then guide diffusion models to leverage its underlying physics knowledge.
To achieve this, we first depict physical concepts with cross-object data (images and corresponding prompts, e.g., ``burning woods'', ``burning match'' and ``burning house'').
In our setting, the physics-related text embedding remain invariant within cross-object prompts since they share the same physics word.
Isometric regularization fine-tunes the text encoder to discover this invariant subspace by enforcing the embeddings of cross-object prompts to achieve inter-similar.
Practically, we set up an anchor prompt ``burning object'' and regularize the variance of the distribution of the distance between the embedding of cross-object prompts and the anchor prompt to be zero, as shown in~\Cref{fig:main_flow}.
Our isometric loss is defined as follows:
\begin{equation}
\label{eq:cross_context_loss}
    \begin{split}
        \Lcal_\text{isometric} = \frac{1}{d} \sum_{i}^d \big( &\Dcal \left(\Ecal_\text{text}(\pB_{\text{a}}), \Ecal_\text{text}(\pB_{\text{i}})\right) \\
        - \frac{1}{d} &\sum_{j}^d \Dcal \left(\Ecal_\text{text}(\pB_{\text{a}}), \Ecal_\text{text}(\pB_{\text{j}})\right) \big)^2 ,
    \end{split}
\end{equation}
where $\pB_\text{i}, i\in\{1,\dots,d\}$ denote $d$ cross-object prompts, $\pB_{\text{a}}$ denotes the anchor prompt, $\Ecal_\text{text}$ is the text encoder of CLIP~\cite{radford2021learning} and $\Dcal(\cdot, \cdot)$ is the Euclidean distance function.

\subsection{Concepts Decoupling}
\label{subsec:concepts_decoupling}

T2I diffusion models often struggle to disentangle multiple given concepts, leading to feature mixture between object and physical concepts.
To address this, we propose decouple loss for the U-Net along with two LoRA modules, one for object concept and one for physical concepts, to encourage the separation of the two independent concepts within diffusion models.
The decouple loss aims at finding two orthogonal spaces for object and physical concept, respectively, in which the learning of each concept is independent, thus eliminating the mixture.
A straightforward approach is to directly 
constrain the weights of the two LoRA modules to be orthogonal, however, this method will force the diffusion models to discard some features for a given concept, resulting in a loss of information.
It is infeasible in physical customization since the physical concepts are more difficult to learn compared with other concepts, which are explicitly depicted by input images.
To address this, we propose to put the orthogonal regularization on the gradients of the two LoRA weights to ensure the independence of different concepts during training.
Compared with regularizing the weights, this smoother approach prevents discarding features and facilitates the diffusion model to choose the features adaptively for different concepts.
Practically, we first calculate the MSE loss for object and physical concepts, denoted as $\Lcal_o$ and $\Lcal_p$:
\begin{align}
\label{eq:mse_loss}
    \Lcal_o &= \Ebb_{\Ecal(\xB_o), \epsilon_o \sim \Ncal(0,1), t}\left[\| \epsilon_o - \Ucal_\theta(\zB_o, \cB_o) \|_2^2 \right] , \\
    \Lcal_p &= \Ebb_{\Ecal(\xB_p), \epsilon_p \sim \Ncal(0,1), t}\left[\| \epsilon_p - \Ucal_\theta(\zB_p, \cB_p) \|_2^2 \right] ,
\end{align}
where $\zB_o$ and $\zB_p$ denote the noisy latent, $\cB_o$ and $\cB_p$ denote the text conditions, $\xB_o$ and $\xB_p$ denote the input images, $\epsilon_o$ and $\epsilon_p$ denote the noises, then decouple loss is defined as follows:
\begin{equation}
\label{eq:decouple_loss}
    \Lcal_\text{decouple} = \frac{\nabla \Lcal_{o}^\top \cdot \nabla \Lcal_{p}}{\left\| \nabla \Lcal_{o} \right\| \cdot \left\| \nabla \Lcal_{p} \right\|},
\end{equation}
where $\nabla \Lcal_{o}$, $\nabla \Lcal_{p}$ are the gradients of the two LoRA modules w.r.t. $\Lcal_o$ and $\Lcal_p$, respectively.
During minimizing $\Lcal_\text{decouple}$, features w.r.t. different concepts are optimized independently without discarding any features, as a result, decoupling the mixture learning of different concepts.
See~\Cref{fig:main_flow} for more details.

\begin{figure*}[t]
    \centering
    \includegraphics[width=\linewidth]{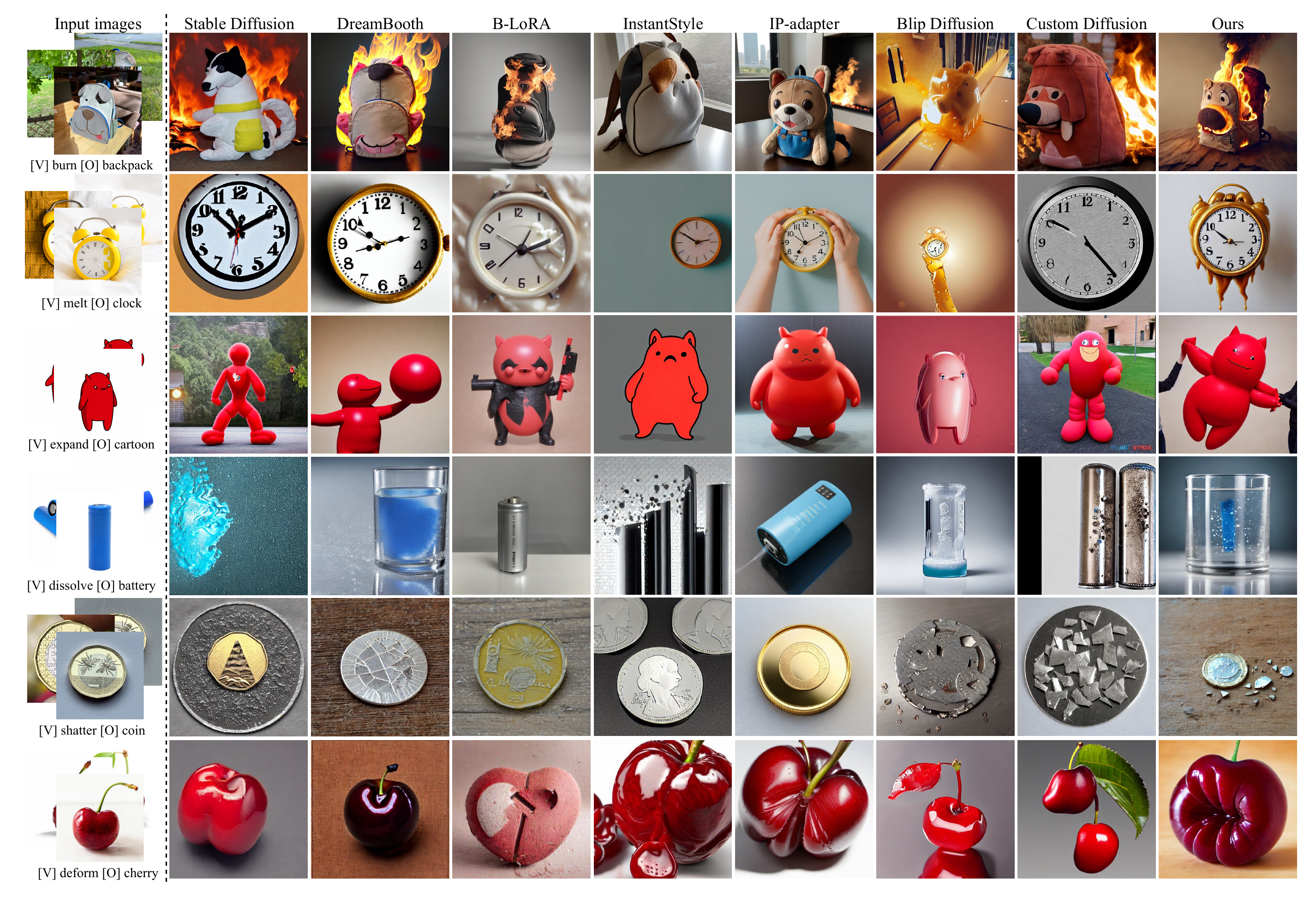}
    \caption{\textbf{Comparison of different methods.} The results show the superior performance of~\methodname~on physical customization.
    }
    \label{fig:comparison_sota}
\end{figure*}

\section{Experiments}

In this section, we first describe our experimental setup in~\Cref{subsec:experimental_setup}. 
We then compare our proposed~\methodname~with SOTA methods in~\Cref{subsec:comparison_with_existing_methods}. 
Next, we present ablation studies in~\Cref{subsec:ablation_studies} to demonstrate the effectiveness of the proposed losses, complemented by additional visualizations in~\Cref{subsec:further_analysis}. 
Finally, we discuss potential applications and limitations of our method in~\Cref{subsec:applications_and_limitations}, providing insights for future development.

\subsection{Experimental Setup}
\label{subsec:experimental_setup}

\paragraph{Implementation details.}
We employ Stable Diffusion v1.5~\cite{Rombach_2022_CVPR} as the foundational model for evaluating our proposed losses.
We also provide results on other foundational models in Appendix, such as Flux.1~\cite{flux2024}.
Our experiments are conducted on an NVIDIA 4080 GPU.
We set the rank of LoRA to be 8.
We use the AdamW~\cite{loshchilov2017decoupled} optimizer with a learning rate of $1 \times 10^{-4}$ and a batch size 4, optimizing for 500 steps.
We set all the loss weights to be 1.
During the inference stage, we set sampling step 50, with a guidance scale of 7.5.

\paragraph{Datasets.}
Our dataset primarily comprises two sets, the object concept set and the physical concept set.
We carefully curate a diverse dataset from previous studies~\cite{ruiz2023dreambooth} and the Internet.
As for the object concept set, each concept contains multiple images depicting a specific object, similar to the dataset released by DreamBooth~\cite{ruiz2023dreambooth}.
As for the physical concept set, each concept contains several images, where each image depicts the same physical concept with different objects.
We carefully construct the physical concept set to include a wide range of real-world transformations, including melting, burning, deforming, dissolving, expanding, fracturing, and shattering.
In total, our dataset contains 25 objects and 6 physical concepts with a total combination of 150 novel concepts.
Dataset samples can be found in Appendix.

\paragraph{Baselines.}
To evaluate the effectiveness of the proposed method, we compare our method with a wide range of popular and SOTA image customization methods, including Blip-diffusion~\cite{li2024blip}, B-LoRA~\cite{frenkel2024implicit}, DreamBooth~\cite{ruiz2023dreambooth}, IP-adapter~\cite{ye2023ip}, Custom Diffusion~\cite{kumari2023multi}, and InstantStyle~\cite{wang2024instantstyle}.
The implementation details of each baseline method can be found in Appendix.

\begin{table}[t]
    \centering
    \begin{adjustbox}{width=\linewidth}
	\begin{threeparttable}
        \begin{tabular}{lcccc}
            \toprule
            Method & CLIP-V $\uparrow$ & CLIP-V-O $\uparrow$ & LLM $\uparrow$ & Average $\uparrow$ \\
            \midrule
            B-LoRA~\cite{frenkel2024implicit} & 0.260 & 0.269 & 0.693 & 0.407 \\
            InstantStyle~\cite{wang2024instantstyle} & 0.233 & 0.299 & 0.716 & 0.416 \\
            IP-adapter~\cite{ye2023ip} & 0.236 & 0.311 & 0.736 & 0.428 \\
            Blip-diffusion~\cite{li2023blip} & 0.264 & 0.273 & 0.780 & 0.439 \\
            DreamBooth~\cite{ruiz2023dreambooth} & 0.265 & 0.293 & 0.787 & 0.448 \\
            Custom Diffusion~\cite{kumari2023multi} & 0.255 & 0.303 & 0.827 & 0.462 \\
            \midrule
            \methodname & \textbf{0.267} & \textbf{0.319} & \textbf{0.840} & \textbf{0.475} \\
            \bottomrule
        \end{tabular}
        \end{threeparttable}
    \end{adjustbox}
    \caption{\textbf{Results comparison with customization SOTA.}~\methodname~achieves overall best performance compared with baselines, demonstrating the superior performance of our proposed method.
    }
    \label{tab:comparison_sota}
\end{table}


\begin{table}[t]
    \centering
    \begin{adjustbox}{width=\linewidth}
	\begin{threeparttable}
        \begin{tabular}{lcccc}
            \toprule
            Method & CLIP-V $\uparrow$ & CLIP-V-O $\uparrow$ & LLM $\uparrow$ & Average $\uparrow$ \\
            \midrule
            \methodname~w/o IL & 0.251 & 0.309 & 0.802 & 0.454 \\
            \methodname~w/o DL & 0.264 & 0.298 & 0.822 & 0.461 \\
            \methodname & \textbf{0.267} & \textbf{0.319} & \textbf{0.840} & \textbf{0.475} \\
            \bottomrule
        \end{tabular}
        \end{threeparttable}
    \end{adjustbox}
    \caption{\textbf{Ablation study results.} w/o IL and w/o DL denote without $\Lcal_\text{decouple}$ and $\Lcal_\text{isometric}$.
    The results demonstrate the effectiveness of both isometric loss and decouple loss.}
    \label{tab:ablation_study}
\end{table}

\begin{table}[t]
    \centering
    \begin{adjustbox}{width=\linewidth}
	\begin{threeparttable}
        \begin{tabular}{lccc}
            \toprule
            Method & Alignment $\uparrow$ & Fidelity $\uparrow$ & Novelty $\uparrow$ \\
            \midrule
            B-LoRA~\cite{frenkel2024implicit} & 15\% & 5\% & 15\% \\
            InstantStyle~\cite{wang2024instantstyle} & 10\% & 10\% & 10\% \\
            IP-adapter~\cite{ye2023ip} & 5\% & 15\% & 5\% \\
            Blip-diffusion~\cite{li2023blip} & 5\% & 10\% & 5\% \\
            DreamBooth~\cite{ruiz2023dreambooth} & 20\% & 20\% & 10\% \\
            Custom Diffusion~\cite{kumari2023multi} & 15\% & 10\% & 20\% \\
            \midrule
            \methodname & \textbf{30\%} & \textbf{30\%} & \textbf{35\%} \\
            \bottomrule
        \end{tabular}
        \end{threeparttable}
    \end{adjustbox}
    \caption{\textbf{User study results.} The results indicate the percentage of people who choose the results of the specific method based on three different criteria.
    The study demonstrates users' preference for our proposed~\methodname, highlighting its effectiveness in generating high-quality, contextually accurate, and innovative images.}
    \label{tab:user_study}
\end{table}

\vspace{-10pt}
\paragraph{Evaluation metrics.}
\label{par:evaluation_metrics}
Our evaluation metrics focus on the relevance between input concepts and generated images. 
We use CLIP~\cite{radford2021learning} to measure similarity between image and text embeddings~\cite{ruiz2023dreambooth}, with higher scores indicating greater relevance, similar to previous studies~\cite{ruiz2023dreambooth, kumari2023multi}.
To assess model's learning of physical concepts, we provide ``A photo of $\{v\}$ object'' as CLIP text input, where $\{v\}$ represents a specific physical concept (e.g., ``melt''), and denote this metric as CLIP-V.
For evaluating the combination of physical concepts and object concepts, we use ``A photo of $\{v\}$ $\{o\}$'' as input, with $\{o\}$ as the specific object, denoted as CLIP-V-O.
Additionally, we use the BLIP-2 VQA model~\cite{li2023blip} to assess relevance by asking: (1) Does the image depict $\{v\}$? (2) Does it depict $\{o\}$? (3) Does it depict $\{v\}$-ing $\{o\}$? 
The metric, denoted as LLM, which is the percentage of ``Yes'' responses across these questions.
For benchmark results, we generate 20 images with different seeds per concept combination (150 total combinations, resulting 3000 images), then select the image with the highest CLIP-V score, then report CLIP-V, CLIP-V-O and LLM scores.



\subsection{Comparison with Existing Methods}
\label{subsec:comparison_with_existing_methods}

\paragraph{Quantitative results.}
CLIP-V measures how well the generated images contain the physical concepts, while CLIP-V-O and LLM measure how well the models compose the object concepts with the physical concepts.
As shown in~\Cref{tab:comparison_sota}, our method achieves the best performance simultaneously on the three metrics, indicating~\methodname~superior performance on activating diffusion models to combine the object concepts with various physical concepts to generate novel concepts.

\begin{figure}[t]
    \centering
    \includegraphics[width=\linewidth]{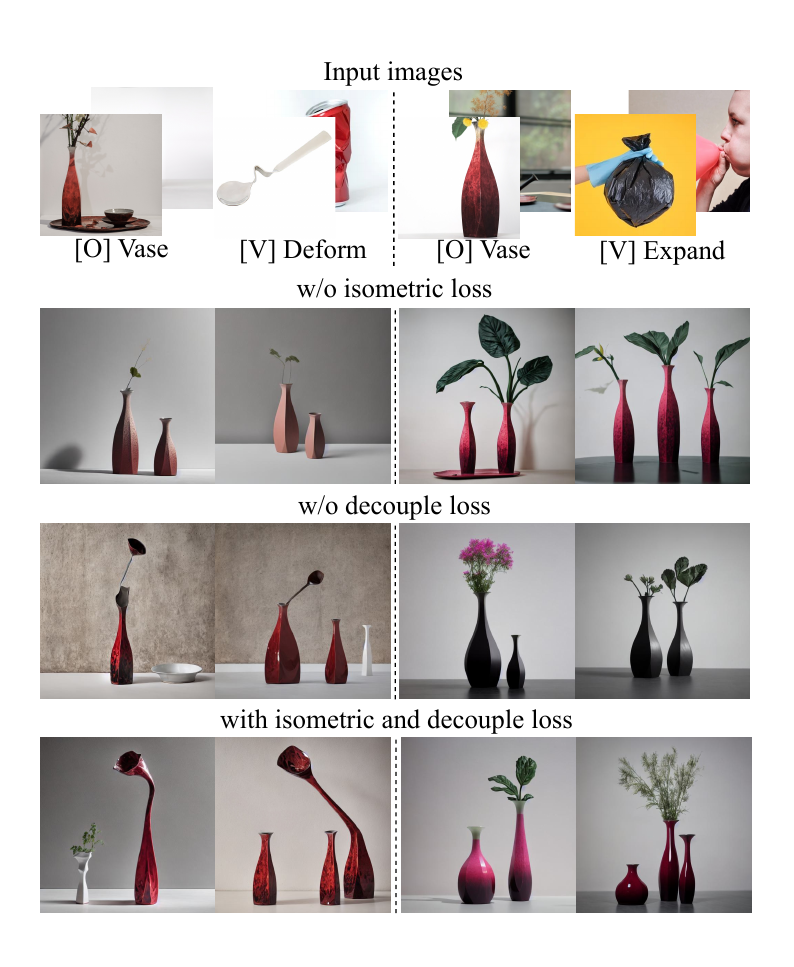}
    \caption{\textbf{Ablation study of the proposed losses.} 
    The results w/o isometric loss fails to perform physical customization, indicating its failure in learning the physical concepts.
    On the other hand, the results without(w/o) decouple loss exhibit pattern leaking, where the model learns to generate the vase combined with the given spoon's shape and the plastic bag's color.}
    \label{fig:ablation_study}
\end{figure}

\paragraph{Qualitative results.}
\Cref{fig:comparison_sota} shows a qualitative comparison of our method against strong baselines, including Stable Diffusion, DreamBooth, IP-adapter, Custom Diffusion, B-LoRA and InstantStyle.
Baseline models struggle to generate meaningful results, falling to provide both fine-grained details and accurate representations of physical transformations.
In contrast, our method outperforms baselines by generating natural and physics-aware images across various combinations of object and physical concepts.
More results can be found in Appendix.

\begin{figure}[t]
    \centering
    \includegraphics[width=\linewidth]{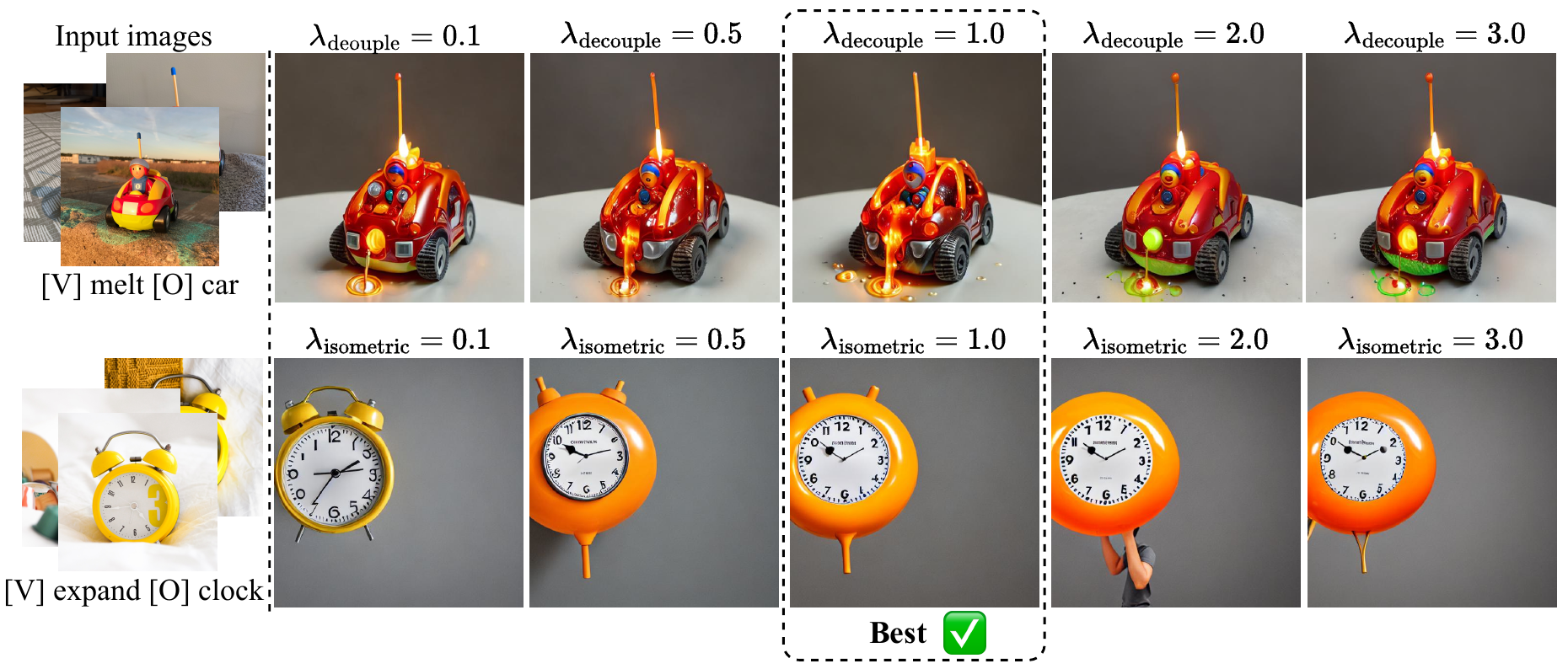}
    \caption{\textbf{Hyper-parameters search.} The results show that setting both $\lambda_\text{isometric}$ and $\lambda_\text{decouple}$ to 1.0 achieves the optimal results.}
    \label{fig:hyper_parameters_search}
\end{figure}

\begin{figure}[t]
    \centering
    \includegraphics[width=0.92\linewidth]{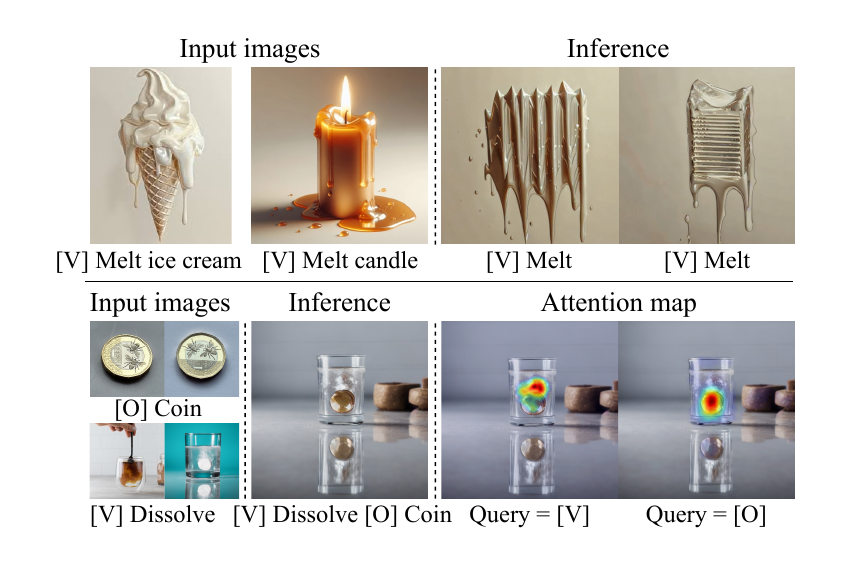}
    \caption{\textbf{Mechanisms of the proposed losses.} Top: The ``[V] Melt'' learned with the isometric loss produces irregular forms with clear melting characteristics, indicating that the isometric loss successfully captures physical knowledge from cross-object data. 
    Bottom: The attention maps for the queries ``[O]'' and ``[V]'' show minimal overlap, demonstrating that the decouple loss effectively disentangles the learning of multiple concepts.}
    \label{fig:daam_attention}
\end{figure}

\paragraph{User study.}
We collected a total of 20 questionnaires for our user study.
We randomly selected 20 concepts from the total 150 combination concepts of our dataset and each participant is required to choose the best image from the generated results of baselines and~\methodname~based on three criteria, resulting in 60 questions.
The criteria are the correspondence between the generated images and the text prompts (Alignment), the fidelity of the generated images to the input objects (Fidelity), and the originality of the generated concepts within the images (Novelty).
The results shown in~\Cref{tab:user_study} indicate users' preference on~\methodname.

\subsection{Ablation Studies}
\label{subsec:ablation_studies}

\Cref{tab:ablation_study} and \Cref{fig:ablation_study} present an ablation study in evaluating the effectiveness of isometric loss and decouple loss, respectively.
More ablation study results can be found in the Appendix.

\begin{figure}[t]
    \centering
    \includegraphics[width=\linewidth]{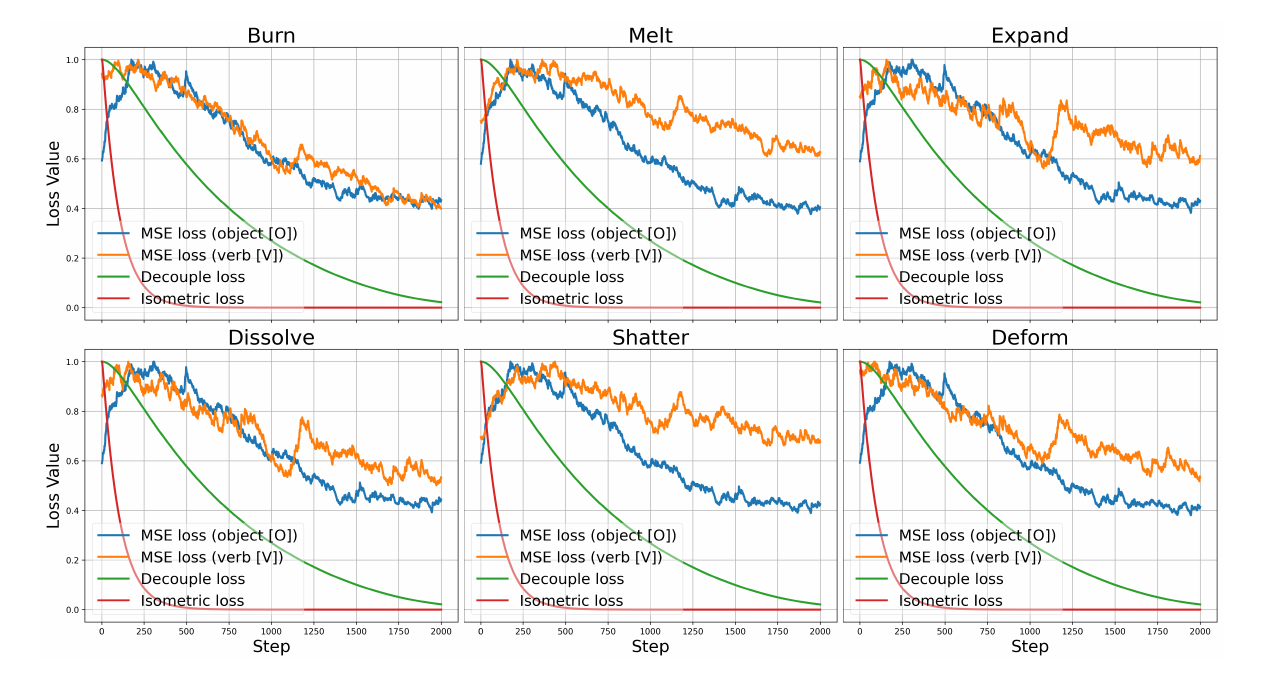}
    \caption{\textbf{Loss curve visualization} under different physical concepts. 
    These curves show the convergence of our proposed losses.
    }
    \label{fig:loss_visualization}
\end{figure}

\paragraph{Isometric regularization.}
\Cref{tab:ablation_study} shows that the performance of~\methodname~without isometric loss degenerates compared with~\methodname.
Specifically,~\methodname~without isometric loss has a relatively much lower CLIP-V score, indicating the failure of the diffusion model to capture the physical knowledge.
Meanwhile, as shown in~\Cref{fig:ablation_study}, when we remove the isometric loss, the images generated solely the object itself without any physical effects. 
Compared with the results generated with isometric loss,~\methodname~can learn and apply the physical concepts on the given object.
These results demonstrate the efficiency of our proposed isometric loss in guiding diffusion models to learn physical concepts from cross-object data.

\paragraph{Concepts decouple regularization.}
\Cref{tab:ablation_study} shows that~\methodname~achieves the best performance compared with~\methodname~without decouple loss.
Specifically,~\methodname~without decouple loss has a relatively lower CLIP-V-O score, indicating diffusion models fail to maintain objects' identity with the existence of pattern leaking.
This is further demonstrated in~\Cref{fig:ablation_study}. Given a deforming spoon,~\methodname~simply generates a vase with a spoon-like head without the regularized by decouple loss, indicating the shape the spoon leaks into the generation.
Moreover, given an expanding plastic bag, the color of the bag leaks into the generation of the vase, resulting in a black vase.
On the contrary, with the help of the decouple loss, our method is able to maintain the style of the input object and independently learns from the physical concept, generating a vase with a deforming neck and a vase with an expanding body shown by the last row in~\Cref{fig:ablation_study}.

\begin{figure}[t]
    \centering
    \includegraphics[width=\linewidth]{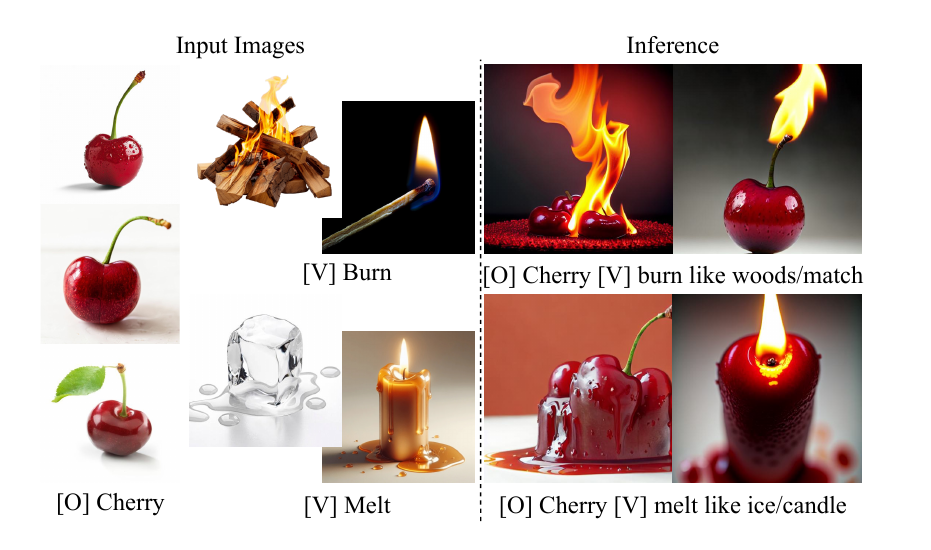}
    \caption{\textbf{Freely control physical concepts.} Given text prompts,~\methodname~is able to control the object to act like the specified object.}
    \label{fig:different_form}
\end{figure}

\subsection{Further Analysis}
\label{subsec:further_analysis}


\paragraph{Hyper-parameters search.}
\Cref{fig:hyper_parameters_search} shows the results of different values of $\lambda_\text{isometric}$ and $\lambda_\text{decouple}$, which are two hyper-parameters indicating the weights of the isometric and decouple loss contribute to the training loss, respectively.
As the figure shows, setting both $\lambda_\text{decouple}$ and $\lambda_\text{isometric}$ to 1 enables to generate optimal results.

\paragraph{Visualizing the mechanisms of the losses.}
To better understand the mechanisms and effectiveness of the proposed losses, we conduct two analyses. 
First, we visualize the generated results using only the physical concept ``[V]'' (without ``[O]'') as the text prompt to examine what the isometric loss enables the diffusion model to learn. 
Second, we visualize the attention maps for queries ``[O]'' ``and [V]'' respectively using DAAM~\cite{tang2022daam} to assess whether the decouple loss effectively prevents the mixture of the two concepts.
As shown in the top row in~\Cref{fig:daam_attention}, the results of the prompt ``[V] Melt'' produce amorphous, unstructured forms with clear melting characteristics but without association to any specific object. 
This suggests that the isometric loss successfully guides the model to learn physics-related features from diverse cross-object data without overfitting to any particular objects.
Meanwhile, in the bottom row, the attention maps demonstrate that the decouple loss effectively separates ``[V]'' and ``[O]'', preventing their entanglement and ensuring each concept is learned independently.

\paragraph{Visualizing the loss curves.}
\Cref{fig:loss_visualization} depicts the loss curves of~\methodname~across six physical concepts: burn, melt, expand, dissolve, shatter, and deform.
These curves demonstrate the convergence of the two proposed losses, highlighting the stability and effectiveness of the proposed framework in learning diverse physical concepts.

\begin{figure}[t]
    \centering
    \includegraphics[width=\linewidth]{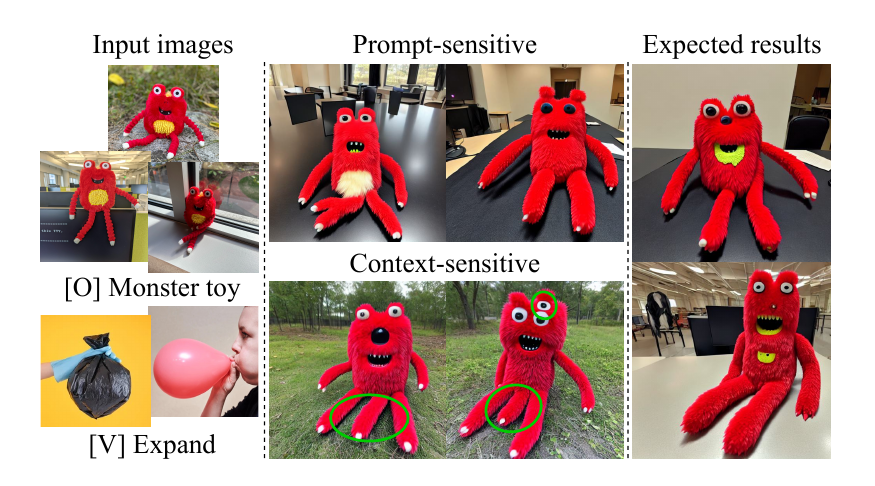}
    \caption{\textbf{Failure cases.} Our method exhibits instability in two scenarios: first, it is sensitive to the quality of input prompts; second, it is affected by the background textures of the input images.}
    \label{fig:failure_cases}
\end{figure}

\subsection{Applications and Limitations}
\label{subsec:applications_and_limitations}


\paragraph{Customization of physical processes.}
As shown in~\Cref{fig:different_form}, our method can flexibly control physical concepts. 
For example, given a pile of burning wood,~\methodname~is able to generate a pile of the specified object burning in a similar manner. 
This adaptability demonstrates that our method not only learns physical knowledge but also freely applies it during generation.

\paragraph{Potential applications.}
Deep learning methods are fundamentally based on the assumption that the test data is drawn from the same distribution as the training data, which is known as the Independent and Identically Distributed assumption. 
As a result, deep neural network methods face significant challenges when encountering Out-of-Distribution (OoD) data.
A key challenge in OoD research is the scarcity of ``corner case'' data~\cite{li2022coda}, which are objects or scenarios rarely seen in the real world.
Even large pre-trained generative models, such as Stable Diffusion~\cite{Rombach_2022_CVPR}, struggle to generate uncommon objects and scenarios effectively. 
To this end, our proposed~\methodname~can serve as a simulator for generating uncommon cases, enabling advancements for a wide range of OoD studies~\cite{wu2023detectbench, zhao2024ood}.
In addition, the synthesized OoD data can serve as a valuable resource for benchmarking the visual reasoning and understanding capabilities~\cite{wu2024vila,chen2022utc}.

\paragraph{Limitations.}
Our method has two main limitations as shown in~\Cref{fig:failure_cases}.
First, it relies on fine-grained text prompts to produce high-quality outputs, which limits its adaptability in scenarios with vague or ambiguous descriptions.
Second, when input images contain complex backgrounds, the diffusion model can be influenced by irrelevant contextual features, leading to inaccuracies in the generated results.
This sensitivity to background complexity may reduce the overall quality of results.

\section{Conclusion}
In this work, we target current limitations in~\task, and propose~\methodname, a framework that freely generates novel concepts and effectively addresses the shortcomings of current diffusion models in understanding physical concepts and performing free combinations between physical concepts and open-world objects.
By introducing isometric and decouple loss functions, our approach enables T2I diffusion models to be physics-aware and accurately learn dynamics, like burning, melting, expanding, dissolving, shattering, and dissolving.
Extensive experiments demonstrate that~\methodname~outperforms existing customization approaches, particularly in its ability to generate controllable and realistic physical transformations, thereby advancing the capabilities of physical customization in generative models.
We believe that this work would act as a foundation for future advancements in~\task, offering potential applications in various scenarios.

{
    \small
    \bibliographystyle{ieeenat_fullname}
    \bibliography{main}
}

\end{document}